\documentclass[runningheads]{llncs}
\usepackage[T1]{fontenc}
\usepackage{graphicx}
\usepackage{amsmath}
\usepackage{amssymb}
\usepackage{multirow}
\usepackage{cite}
\usepackage{hyperref}
\usepackage[table,xcdraw]{xcolor}
\usepackage{pifont}
\usepackage[]{algorithm2e}
\newcommand{\R}{\mathbb{R}}

\newcommand{\Dthree}{$\mathcal{D}_{300}$}
\newcommand{\Dsix}{$\mathcal{D}_{600}$}
\newcommand{\Dnine}{$\mathcal{D}_{900}$}
\newcommand{\Dtwelve}{$\mathcal{D}_{1200}$}

\DeclareMathOperator{\E}{\mathbb{E}}

\SetKwInput{kwCompute}{Compute}
\SetKwInput{kwDefine}{Define}
\SetKwInput{kwInput}{Input}

\usepackage{color}
\begin{document}
\title{Learning shape distributions from large databases of healthy organs: applications to zero-shot and few-shot abnormal pancreas detection}

\titlerunning{Learning shape distributions from large databases of healthy organs}
%
\author{Rebeca V\'etil\inst{1,2,*} \and
Cl\'ement Abi-Nader\inst{2} \and
Alexandre B\^one\inst{2} \and
Marie-Pierre Vullierme\inst{3} \and
Marc-Michel Roh\'e\inst{2} \and
Pietro Gori\inst{1} \and
Isabelle Bloch\inst{1,4}
}


%
\authorrunning{V\'etil at al.}
%
\institute{LTCI, T\'el\'ecom Paris, Institut Polytechnique de Paris, France \and
Guerbet Research, Villepinte, France \and
Department of Radiology, Hospital of Annecy-Genevois, Universit\'e de Paris, France \and
Sorbonne Universit\'e, CNRS, LIP6, Paris, France \\
\email{rebeca.vetil@guerbet.com}}
\maketitle              
\begin{abstract}
We propose a scalable and data-driven approach to learn shape distributions from large databases of healthy organs. To do so, volumetric segmentation masks are embedded into a common probabilistic shape space that is learned with a variational auto-encoding network. The resulting latent shape representations are leveraged to derive zero-shot and few-shot methods for abnormal shape detection. The proposed distribution learning approach is illustrated on a large database of 1200 healthy pancreas shapes. Downstream qualitative and quantitative experiments are conducted on a separate test set of 224 pancreas from patients with mixed conditions. The abnormal pancreas detection AUC reached up to $65.41\%$ in the zero-shot configuration, and $78.97\%$ in the few-shot configuration with as few as $15$ abnormal examples, outperforming a baseline approach based on the sole volume.
\end{abstract}
\section{Introduction}
Anatomical alterations of organs such as the brain or the pancreas may be informative of functional impairments. For instance, hippocampal atrophy and duct dilatation are well-known markers of Alzheimer's disease and pancreatic ductal adenocarcinoma, respectively~\cite{fox1996presymptomatic, liu2019jointshape}. In these examples, quantifying anatomical differences bears therefore a great potential for 
determining the patient's clinical status, anticipating its future progression or regression, and supporting the treatment planning. \let\thefootnote\relax\footnotetext{*Corresponding author: \email{rebeca.vetil@guerbet.com}}
\\ \indent Since the seminal work of Thompson~\cite{thompson1917growth}, the computational anatomy literature proposed several Statistical Shape Modeling (SSM) approaches, which embed geometrical shapes into metric spaces where notions of distance and difference can be defined and quantified~\cite{kendall1984shape, christensen1996deformable, beg2005computing}. Taking advantage of these representations, statistical shape models were then proposed to perform group analyses of shape collections. In particular, atlas models~\cite{pennec2006intrinsic} learn geometrical distributions in terms of an ``average'' representative shape and associated variability, generalizing the Euclidean mean-variance analysis. In medical imaging, learning atlases from healthy examples allows for the definition of normative models for anatomical structures or organs, such as brain MRIs or subcortical regions segmented from neuroimaging data \cite{zhang2013bayesian, gori2017bayesian}, thus providing a natural framework for the detection of abnormal anatomies. \\ \indent In practice, leveraging an atlas model to compute the likelihood of a given shape to belong to the underlying distribution either requires to identify landmarks~\cite{cootes1995_ASM}, or to solve a registration problem~\cite{bone2018_deformetrica}. To circumvent the computational cost of this shape embedding operation, the authors in~\cite{yang2017quicksilver} proposed to train an encoder network to predict registration parameters from image pairs. In~\cite{dalca2018unsupervised, krebs2019learning}, the authors built on this idea and used the variational autoencoder (VAE) of~\cite{kingma2014} to learn the embedding space jointly with the atlas model, instead of relying on pre-determined parametrization strategies. However, the structure of the decoding network remained constrained by hyperparameter-rich topological assumptions, enforced via costly smoothing and numerical integration operators from a computational point of view. \\ \indent Alternative approaches proposed to drop topological hypotheses by relying on variations of the AE or its variational counterparts~\cite{kingma2014} to learn normative models that are subsequently used to perform Anomaly Detection (AD). These methods compress and reconstruct images of healthy subjects to capture a normative model of organs~\cite{baur2021autoencoders, Zimmerer2019UnsupervisedAL}. Yet, they are usually applied on the raw imaging data, thus they entail the risk of extracting features related to the intensity distribution of a dataset which are not necessarily specific to the organ anatomy. Therefore, regularization constraints~\cite{baur2021autoencoders, chen2018unsupervised} are introduced to improve the detection performances compared to the vanilla AE. To further reduce the overfitting risk, these methods artificially increase the dataset size by working on 2D slices.
\\ \indent Given this context, we propose a VAE-based method to learn a normative model of organ shape that can subsequently be used to detect anomalies, thus bridging the gap between SSM and AD models. Although SSM methods with explicit modeling constraints proved effective to learn relevant shape spaces from relatively small collections of high-dimensional data, we propose to further reduce the set of underlying hypotheses and leave the decoding network unconstrained in its architecture.
With the objective to learn normative shape models from collections of healthy organs, we argue that sufficiently large databases of relevant medical images can be constructed by pooling together different data sources, see~\cite{dufumier2021contrastive} for instance. To reduce the risk of overfitting and focus on the anatomy of organs, the VAE is learned from 3D binary segmentation masks and is coupled with a shape-preserving data augmentation strategy consisting of translations, rotations and scalings. An approach to study and visualize group differences is also proposed. \\ \indent Section~\ref{sec:method} details the proposed method, which is then illustrated on a pancreas shape problem in Section~\ref{sec:results}. Section~\ref{sec:conclusion} discusses the results and concludes. 
\section{Methods}
\label{sec:method}

\textbf{Modeling organ shape.}
We consider an image acquired via a standard imaging technique. For a given organ in the image, its anatomy can be represented by a binary segmentation mask $\mathbf{X} = \{ x_i, i=1...d \}$ with $x_i \in \{0, 1\}$ and $d$ the number of voxels in the image. We are interested in studying the shape of this organ, and assume that it is characterized by a set of underlying properties that can be extracted from the segmentation mask. Therefore, we hypothesize the following generative process for the segmentation mask:
\begin{align}
    p_{\theta}(\mathbf{X} \mid \mathbf{z}) & = \prod_{i=1}^{d}f_{\theta}(\mathbf{z})_{i}^{x_i}(1-f_{\theta}(\mathbf{z})_{i})^{1-x_i}
\end{align}
where $0^0 = 1$ by convention, and $\mathbf{z}$ is a latent variable generated from a prior distribution $p(\mathbf{z})$. This latent variable provides a low-dimensional representation of the segmentation mask embedding its main shape features. The function $f_{\theta}$ is a non-linear function mapping $\mathbf{z}$ to a predicted probabilistic segmentation mask. 

We are interested in inferring the parameters $\theta$ of the generative process, as well as approximating the posterior distribution of the latent variable $\mathbf{z}$ given a segmentation mask $\mathbf{X}$. We rely on the VAE framework \cite{kingma2014} to estimate the model parameters. Hence, we assume that $p(\mathbf{z})$ is a multivariate Gaussian with zero mean and identity covariance. We also introduce the approximate posterior distribution $q_{\phi}(\mathbf{z} \mid \mathbf{X})$ parameterized by $\phi$, and optimize a lower bound $\mathcal{L}$ of the marginal log-likelihood, which can be written for the segmentation mask $\mathbf{X}^{p}$ of a subject $p$ as: 
\begin{align}
    \mathcal{L} = \E_{q_{\phi}(\mathbf{z} \mid \mathbf{X}^{p})}[\log p_{\theta}(\mathbf{X}^{p} \mid \mathbf{z})] - KL[q_{\phi}(\mathbf{z} \mid \mathbf{X}^{p}) \mid p(\mathbf{z})],
\end{align}
where $q_{\phi}(\mathbf{z} \mid \mathbf{X}^{p})$ follows a Gaussian distribution $ \mathcal{N}(\mu_{\phi}(\mathbf{X}^{p}), \sigma^{2}_{\phi}(\mathbf{X}^{p})\mathbf{I})$ with $\mathbf{I}$ the identity matrix, and $KL$ is the Kullback-Leibler divergence. 

To capture shape features, we rely on a convolutional network and adopt the U-Net \cite{ronneberger2015} encoder-decoder architecture without skip connections$^{\ast}$\let\thefootnote\relax\footnotetext{$^{\ast}$Code available at \href{https://github.com/rebeca-vetil/HealthyShapeVAE}{https://github.com/rebeca-vetil/HealthyShapeVAE}.}. In practice, the number of convolutional layers and the convolutional blocks are automatically inferred thanks to the nnU-Net self-configuring procedure~\cite{isensee2021nnunet} (see Section A in the supplementary material for details). Due to this encoder-decoder architecture, the segmentation masks are progressively down-sampled to obtain low-resolution feature maps which are mapped through a linear transformation to the latent variable $\mathbf{z}$. The latent code is subsequently decoded by a symmetric path to reconstruct the original masks.

\indent\textbf{Anomaly detection.}
\label{ssec:ab_detection}
We propose to learn a normative model of organ shapes by applying the VAE framework previously presented on the segmentation masks of a large cohort of $N$ healthy patients, allowing the model to capture a low-dimensional embedding characteristic of a normal organ anatomy. In addition, we use a data augmentation procedure consisting of random translations, rotations and scalings, in order to be invariant to these transformations and force the network to extract shape features. Based on this learned model, we propose two approaches to perform Anomaly Detection (AD) by leveraging the latent representation of normal organ shapes.

\indent\textbf{Zero-shot learning method.} After training, the recognition model $q_{\phi}(\mathbf{z} \mid \mathbf{X})$ can be used to project the segmentation maps $\mathbf{X}^{p}$ of the cohort of healthy subjects and obtain an empirical distribution of normal shapes in the latent space. We rely on this low-dimensional distribution of normality to detect abnormal shapes. To do so, we compute the mean of the healthy subjects projection, and define abnormality through the L2 distance to this mean latent representation.

\indent\textbf{Few-shot learning method.} Another approach is to classify normal and abnormal shapes based on their low-dimensional representations. In practice, we project the segmentation maps from a set of healthy and pathological subjects in the latent space using the recognition model $q_{\phi}(\mathbf{z} \mid \mathbf{X})$. Therefore, we obtain for all these subjects a set of low-dimensional organ shape features that we can use to learn any type of classifier (\textit{e.g.,} linear SVM).

\indent\textbf{Studying organ shapes differences.}
\label{ssec:organ_deformation}
Our framework can also be used to study organ differences between groups. Let us consider a set of healthy and pathological subjects, as well as their segmentation masks. Based on the recognition model $q_{\phi}(\mathbf{z} \mid \mathbf{X})$, we can compute the average of the subject's latent projection for each group, denoted by $\mathbf{z}_{normal}$ and $\mathbf{z}_{abnormal}$, respectively. We consider the line of equation $(1-t) \times \mathbf{z}_{normal} + t\times \mathbf{z}_{abnormal}$ with $t \in \R$. When moving along this line with increasing values of $t$, we progress from a healthy mean latent shape representation to a pathological one, and can reconstruct the corresponding segmentation mask using the probabilistic decoder $p_{\theta}(\mathbf{X} \mid \mathbf{z})$.

\section{Experiments}
\label{sec:results}
In this section, we applied our method in the case of the pancreas. A normative model of pancreas shape was learned on a large cohort of healthy subjects, and was then leveraged for AD on an independent test cohort. Several configurations were proposed to assess the model performances, including the impact of the number of training subjects and of the latent space dimensionality on the AD performances. Detection with the few-shot learning method was performed using Support Vector Machine (SVM). Finally, we showed how the proposed framework can be used to visualize differences between the healthy and pathological pancreas.

\indent\textbf{Training.} The training dataset $\mathcal{D}^{Train}$ was created from our private cohort containing $2606$ abdominal Portal CT scans of patients with potential liver cancer. To ensure the healthy condition and shape of the pancreas, several exclusion criteria were applied (see Section B in the supplementary material). Finally, $1 200$ portal CT scans were retained. To explore the influence of the number of samples seen during training, subsets $\mathcal{D}_{N}$ with a growing number of subjects were created (see Table $1$ in the supplementary material). For each $\mathcal{D}_{N}$, $80\%$ and $20\%$ of the samples were used for training and validation, respectively. Splitting was done such that the pancreas volume distribution was balanced across the splits.

\begin{figure}[t]
\centering
\includegraphics[width=0.4\textwidth]{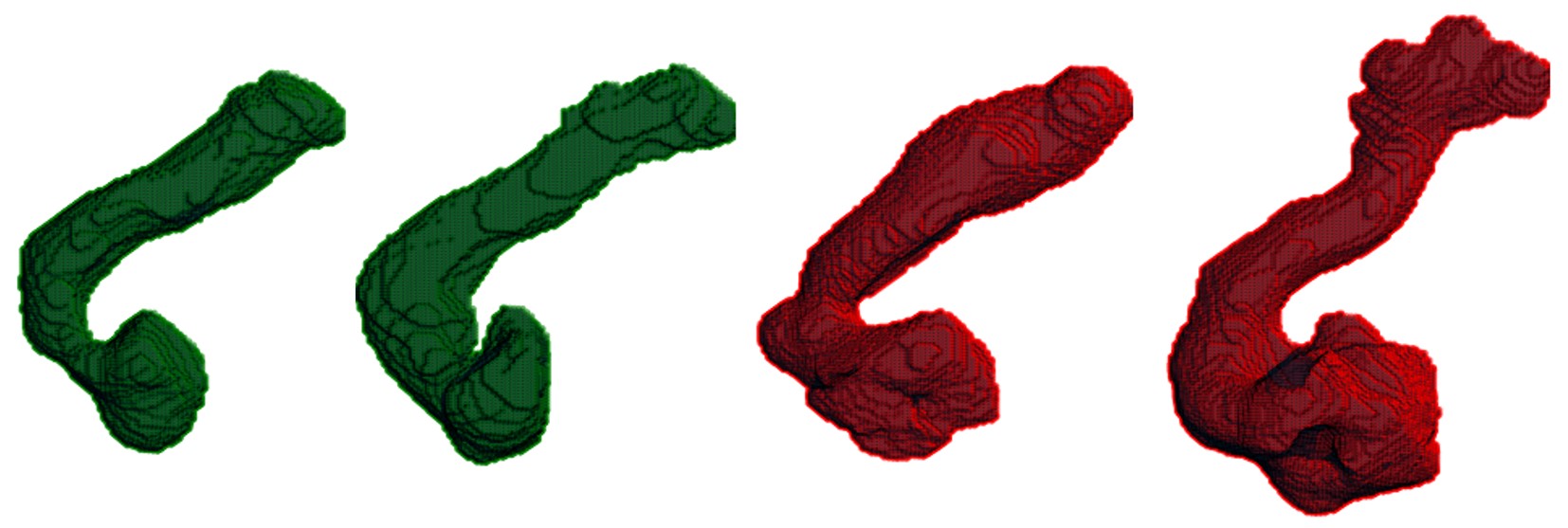}
\caption{\textbf{Examples of normal and abnormal pancreas shapes from $\mathcal{D}^{Test}$}. Green and red figures are examples taken from $\mathcal{D}_{normal}^{Test}$ and $\mathcal{D}_{abnormal}^{Test}$, respectively.} \label{fig:examples}
\end{figure}

\indent\textbf{Testing.} The test database $\mathcal{D}^{Test}$ was obtained by combining two datasets: i) a private dataset $\mathcal{D}_{abnormal}^{Test}$ containing $144$ cases diagnosed with pancreatic cancer, and for whom the pancreas shape was evaluated as abnormal by an expert radiologist; ii) a public dataset $\mathcal{D}_{normal}^{Test}$ from The Cancer Imaging Archive (TCIA) containing $80$ CT scans \cite{TCIA} of patients who neither had abdominal pathologies nor pancreatic lesions, and for whom the assumption of normal pancreas shape held. Centers, machines and protocols differed among the three datasets. Examples of normal and abnormal shapes can be seen in Figure \ref{fig:examples}.

\indent\textbf{Preprocessing.} 
The first step consisted in obtaining the pancreas segmentation masks. For the public dataset $\mathcal{D}_{normal}^{Test}$, we used the reference pancreas segmentation masks provided by TCIA. For $\mathcal{D}_{normal}^{Train}$ and $\mathcal{D}_{abnormal}^{Test}$, the masks were obtained semi-automatically using an in-house segmentation algorithm derived from the nnU-Net, and validated by a radiologist with 25 years of expertise in abdominal imaging. Finally, all the masks were resampled to $1 \times 1 \times 2$ mm$^3$ in $(x, y, z)$ directions, and centered in a volume of size $192 \times 128 \times 64$ voxels.

\indent\textbf{Zero-shot AD.}
We trained our model on the different datasets $\mathcal{D}^{Train}_{N}$, with a growing number of latent dimensions $L$ ranging from $16$ to $1024$ (denoted by $L_{16} ... L_{1024}$). For each experiment, we applied the zero-shot AD procedure, as previously explained, on $\mathcal{D}^{Test}$. We report the Area Under the Curve (AUC), in \%, in Table \ref{tab:unsupervised}. Increasing the dimension of the latent space $L$ improved the classification performances on each dataset $\mathcal{D}_{N}$. Moreover, for each dataset size the best result was consistently obtained when $L$ was set at the maximum value $L_{1024}$. We also observed that the effect of the latent space dimension on the performances seemed to attenuate as the dataset size increased. Indeed, we observed that when going from $L_{16}$ to $L_{1024}$, the mean AUC for \Dthree, \Dsix, \Dnine, \Dtwelve\ improved by $10.7$, $5.9$, $4.0$ and $3.1$ points, respectively. Regarding the effect of the database size, we observed that increasing the training set size seemed to globally improve the AUC scores. For instance, going from \Dthree\ to \Dsix\ increased the classification performances for all the experiments, particularly for $L_{16}$ which gained $9.3$ points. This beneficial effect of both larger training sets and latent dimension was also observed on the Dice score between the original and reconstructed segmentation masks (see Table $2$ in the supplementary material). Thus, for the following experiments, we chose the model trained on \Dtwelve\ with a latent dimension $L_{1024}$ as it gave the best results in terms of AUC and Dice scores.

\begin{table}[h]
\centering
\caption{\textbf{Results for zero-shot AD.} For each experiment, corresponding to a specific training size $\mathcal{D}$ and latent space dimension $L$, we report the mean and standard deviation of AUC scores (in \%) obtained by bootstrapping with $10000$ repetitions. Best results by line are \underline{underlined} and by column are in \textbf{bold}.}
\label{tab:unsupervised}
\begin{tabular}{c||c|c|c|c}
 & ${L_{16}}$ & ${L_{64}}$ & ${L_{256}}$ & ${L_{1024}}$ \\ \hline
\textbf{\Dthree} & 51.51 \scriptsize{$\pm$0.37} & 59.08 \scriptsize{$\pm$0.37} & 62.16 \scriptsize{$\pm$0.37} & \underline{62.17} \scriptsize{$\pm$0.37} \\
\textbf{\Dsix} & 59.24 \scriptsize{$\pm$0.38} & 60.97 \scriptsize{$\pm$0.36} & \textbf{64.32} \scriptsize{$\pm$0.36} & \underline{65.11} \scriptsize{$\pm$0.36} \\ 
\textbf{\Dnine} & 60.77 \scriptsize{$\pm$0.37} & \textbf{62.64} \scriptsize{$\pm$0.37} & 64.04 \scriptsize{$\pm$0.36} & \underline{64.81} \scriptsize{$\pm$0.36} \\ 
\textbf{\Dtwelve} & \textbf{62.28} \scriptsize{$\pm$0.36} & 61.74 \scriptsize{$\pm$0.37}& 62.58 \scriptsize{$\pm$0.37} & \underline{\textbf{65.41}} \scriptsize{$\pm$0.36} \\
\end{tabular}%
\end{table}

\begin{figure}
\includegraphics[width=\textwidth]{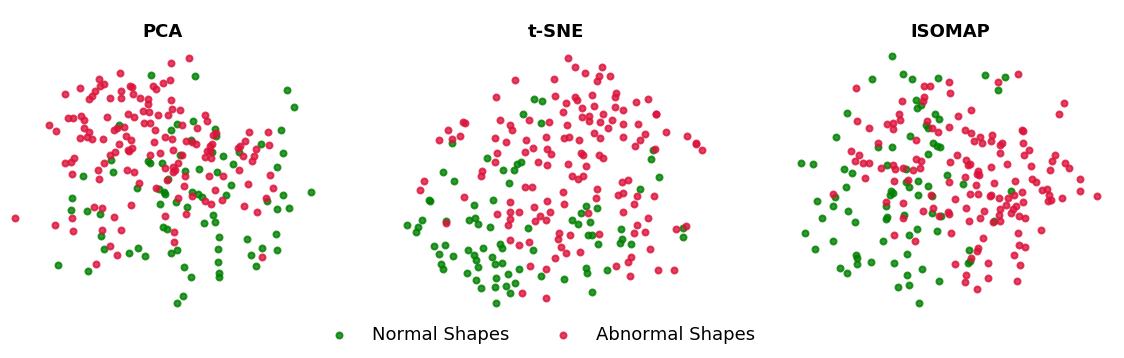}
\caption{\textbf{2D reduction of the latent representation of the test subjects.} The $80$ samples from $\mathcal{D}_{normal}^{Test}$ are in green, and the $144$ samples from $\mathcal{D}_{abnormal}^{Test}$ are in red.} \label{fig:proj}
\end{figure}

To visualize the separation between normal and abnormal shapes, we projected each subject from $\mathcal{D}^{test}$ using the recognition model $q_{\phi}(\mathbf{z} | \mathbf{X})$. Based on the subjects' latent representation, we applied three dimensionality reduction techniques, namely Principal Components Analysis (PCA), t-distributed Stochastic Neighbor Embedding (t-SNE) and Isomap. Results are displayed in Figure \ref{fig:proj}, on which each point represents the latent projection of a test subject reduced on a 2D plane. We observed that, independently of the projection technique, normal and abnormal shapes tended to be separated in two different clusters. 

\indent\textbf{Few-shot AD.} We trained a linear SVM classifier on the latent representation of $\mathcal{D}^{Test}$ with stratified k-fold cross-validation. We varied the number $k$ of folds to test the performances of the classifier depending on the train/test samples ratio. Experiments ranged from a $0.05$ train/test ratio to a leave-one-out cross-validation and are presented in Table \ref{tab:supervised-fewshot}. We noticed that using only $8$ healthy and $15$ abnormal training samples increased the performance to $78.9\%$. We also observed that the AUC scores and the balanced accuracy increased with the number of training samples, reaching a maximum of $91.1\%$ and $83.2\%$ respectively in the leave-one-out configuration. 

\begin{table}[h]
\centering
\caption{\textbf{Results for few-shot AD.} For each experiment, we indicate the number of training samples, as well as the number of abnormal samples (in brackets). We report the means and standard deviations for AUC (in \%) and balanced accuracy (in \%), obtained by bootstrapping with $10000$ repetitions.}
\label{tab:supervised-fewshot}
\begin{tabular}{c||c|c|c|c|c}
\textbf{Train/Test ratio} & \textbf{0.05} & \textbf{0.11} & \textbf{0.25} & \textbf{1} & \textbf{223} \\
\textit{Number of training samples} & \textit{12 (8)} & \textit{23 (15)} & \textit{45 (29)} & \textit{112 (72)} & \textit{223 (144)} \\ \hline
\textbf{AUC} & 66.02 \scriptsize{$\pm$0.02} & 78.97 \scriptsize{$\pm$0.03} & 81.87 \scriptsize{$\pm$0.07} & 86.95 \scriptsize{$\pm$0.23} & 91.18 \scriptsize{$\pm$0.19} \\
\textbf{Balanced Accuracy} & 67.78 \scriptsize{$\pm$0.03} & 70.88 \scriptsize{$\pm$0.05} & 73.96 \scriptsize{$\pm$0.10} & 75.49 \scriptsize{$\pm$0.37} & 83.26 \scriptsize{$\pm$0.34}
\end{tabular}%
\end{table}

\indent\textbf{Comparison with a baseline method.} We compared our approach with a baseline method classifying shapes based on their volume. We applied this method on $\mathcal{D}^{Test}$ with bootstrap sampling and obtained an average AUC of $51\%$ with a $95\%$ confidence interval of $[49.9; 51.7]$, below the maximum AUC scores of $65.4\%$ and $91.1\%$ previously reported in the zero-shot and few-shot cases, respectively. We also compared the proposed method to two SSM methods: active shape models (ASM)~\cite{cootes1995_ASM} and Large Deformation Diffeomorphic Metric Mapping (LDDMM) using the Deformetrica software~\cite{bone2018_deformetrica}. Details and results, reported in Section D in the supplementary material, show that our approach outperforms these two state-of-the-art methods in both zero and few-shot configurations.

\begin{figure}[b]
\includegraphics[width=\textwidth]{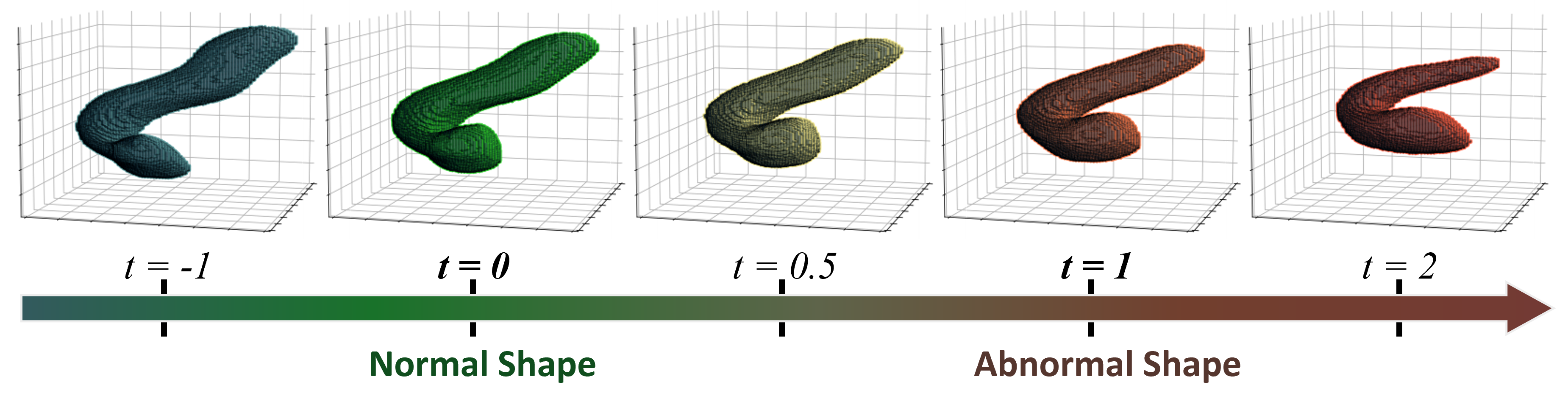}
\caption{\textbf{Generated pancreas shapes.} Pancreas shapes generated by decoding latent representations lying on the line of equation$(1-t) \times \mathbf{z}_{normal} + t \times \mathbf{z}_{abnormal}$.} \label{fig:progression}
\end{figure}

\indent\textbf{Studying pancreas shapes differences between groups.} To model differences in the pancreas shape between healthy and pathological groups, we applied the procedure presented in Section \ref{ssec:organ_deformation} on the subjects from $\mathcal{D}^{Test}_{normal}$ and $\mathcal{D}^{Test}_{abnormal}$. Figure \ref{fig:progression} shows the pancreas shapes obtained for different values of~$t$. When going from a healthy towards a pathological latent representation, we observed a shrinkage of the shape in the body for the generated pancreas.

\section{Discussion and conclusion}
\label{sec:conclusion}
We presented a method based on a VAE to learn a normative model of organ shape. We hypothesized that such a model could be learned from large databases of healthy subjects. The method was applied in the case of the pancreas, for which morphological changes can be a marker of disease. 

We empirically observed that large training sets and latent dimensions were beneficial to the model in terms of AD performances. Our results also demonstrated that the model captured features that distinguished between normal and abnormal shapes in the latent space, as illustrated in Figure \ref{fig:proj}. From a quantitative point of view, we observed in the zero-shot case - \textit{i.e.} without supervision - that the best model obtained an AUC score of $65.41 \pm 0.36 \%$, which significantly outperformed a naive model classifying shapes based on their volume. In the few-shot experiments, we obtained a mean AUC score of $77.4\%$ by training a SVM with only $8$ healthy subjects and $15$ pathological subjects. These findings highlight the discriminating properties of the latent normative model of pancreas shape estimated by our model. 
Moreover, classification performances reached up to $91.1\%$ AUC and $83.2 \%$ balanced accuracy when training the classifier on $223$ samples in a leave-one-out fashion. These results are in line with \cite{liu2019jointshape}, where the authors reported a balanced accuracy of $85.2\%$ on their private dataset. Yet, our approach differs from theirs by its paradigm. Instead of training a supervised model for joint shape representation and classification, we propose to learn a normative model of shape. The advantage of this approach is that it does not require different types of patients to be trained but solely a database of healthy subjects. Moreover, it can be used in an unsupervised manner (\textit{cf.} zero-shot) or with few labeled data (\textit{cf.} few-shot), with good performances in both cases. 

Finally, we also showed that our framework could be used to study and visualize the morphological differences between the organ shape of different clinical groups, based on an exploration of the latent space. The anatomical changes observed in Figure \ref{fig:progression} seemed to concur with clinical evidence as the shrinkage suggests partial parenchymal atrophy \cite{yamao2020partial}. This hypothesis would require further medical evaluation, and could be the subject of a proper clinical validation.

\indent\textbf{Acknowledgments} This work was partly funded by a CIFRE grant from ANRT \# $2020/1448$.

%
%
%
\newpage
\bibliographystyle{splncs04}
\bibliography{biblio}

\begin{thebibliography}{10}
\providecommand{\url}[1]{\texttt{#1}}
\providecommand{\urlprefix}{URL }
\providecommand{\doi}[1]{https://doi.org/#1}

\bibitem{baur2021autoencoders}
Baur, C., Denner, S., Wiestler, B., Navab, N., Albarqouni, S.: Autoencoders for
  unsupervised anomaly segmentation in brain {MR} images: a comparative study.
  Medical Image Analysis  \textbf{69},  101952 (2021)

\bibitem{beg2005computing}
Beg, M., Miller, M., Trouv{\'e}, A., Younes, L.: Computing large deformation
  metric mappings via geodesic flows of diffeomorphisms. IJCV  \textbf{61}(2),
  139--157 (2005)

\bibitem{bone2018_deformetrica}
B\^one, A., Louis, M., Martin, B., Stanley, D.: Deformetrica 4: An open-source
  software for statistical shape analysis. In: International Conference on
  Medical Image Computing and Computer-Assisted Intervention. pp. 3--13.
  Springer (2018)

\bibitem{chen2018unsupervised}
Chen, X., Konukoglu, E.: Unsupervised detection of lesions in brain {MRI} using
  constrained adversarial auto-encoders. In: MIDL (2018)

\bibitem{christensen1996deformable}
Christensen, G.E., Rabbitt, R.D., Miller, M.I.: Deformable templates using
  large deformation kinematics. IEEE Transactions on Image Processing
  \textbf{5}(10),  1435--1447 (1996)

\bibitem{cootes1995_ASM}
Cootes, T.F., Taylor, C.J., Graham, J.: Active shape models-their training and
  application. Computer Vision and Image Understanding  \textbf{61}(1),  38--50
  (1995)

\bibitem{dalca2018unsupervised}
Dalca, A.V., Balakrishnan, G., Guttag, J., Sabuncu, M.R.: Unsupervised learning
  for fast probabilistic diffeomorphic registration. In: International
  Conference on Medical Image Computing and Computer-Assisted Intervention. pp.
  729--738. Springer (2018)

\bibitem{dufumier2021contrastive}
Dufumier, B., Gori, P., Victor, J., Grigis, A., Wessa, M., Brambilla, P.,
  Favre, P., Polosan, M., Mcdonald, C., Piguet, C.M., et~al.: Contrastive
  learning with continuous proxy meta-data for {3D} {MRI} classification. In:
  International Conference on Medical Image Computing and Computer-Assisted
  Intervention. pp. 58--68. Springer (2021)

\bibitem{fox1996presymptomatic}
Fox, N., Warrington, E., Freeborough, P., Hartikainen, P., Kennedy, A.,
  Stevens, J., Rossor, M.N.: Presymptomatic hippocampal atrophy in
  {Alzheimer's} disease: a longitudinal {MRI} study. Brain  \textbf{119}(6),
  2001--2007 (1996)

\bibitem{gori2017bayesian}
Gori, P., Colliot, O., Marrakchi-Kacem, L., Worbe, Y., Poupon, C., Hartmann,
  A., Ayache, N., Durrleman, S.: {A Bayesian Framework for Joint Morphometry of
  Surface and Curve meshes in Multi-Object Complexes}. {Medical Image Analysis}
   \textbf{35},  458--474 (2017)

\bibitem{isensee2021nnunet}
Isensee, F., et~al.: nn{U-N}et: a self-configuring method for deep
  learning-based biomedical image segmentation. Nature Methods  \textbf{18}(2),
   203--211 (2021)

\bibitem{kendall1984shape}
Kendall, D.G.: Shape manifolds, {P}rocrustean metrics, and complex projective
  spaces. Bulletin of the London Mathematical Society  \textbf{16}(2),  81--121
  (1984)

\bibitem{kingma2014}
Kingma, D.P., Welling, M.: {Auto-Encoding Variational Bayes}. In: 2nd
  International Conference on Learning Representations, {ICLR} 2014, Banff, AB,
  Canada (2014)

\bibitem{krebs2019learning}
Krebs, J., Delingette, H., Mailh{\'e}, B., Ayache, N., Mansi, T.: Learning a
  probabilistic model for diffeomorphic registration. IEEE Transactions on
  Medical Imaging  \textbf{38}(9),  2165--2176 (2019)

\bibitem{liu2019jointshape}
Liu, F., Xie, L., Xia, Y., Fishman, E., Yuille, A.: Joint shape representation
  and classification for detecting {PDAC}. In: International Workshop on
  Machine Learning in Medical Imaging. pp. 212--220. Springer (2019)

\bibitem{pennec2006intrinsic}
Pennec, X.: Intrinsic statistics on {R}iemannian manifolds: Basic tools for
  geometric measurements. Journal of Mathematical Imaging and Vision
  \textbf{25}(1),  127--154 (2006)

\bibitem{ronneberger2015}
Ronneberger, O., Fischer, P., Brox, T.: {U-Net}: Convolutional networks for
  biomedical image segmentation. In: International Conference on Medical Image
  Computing and Computer-Assisted Intervention. pp. 234--241. Springer (2015)

\bibitem{TCIA}
Roth, H., et~al.: Data from pancreas-{CT}. {The Cancer Imaging Archive} (2016).
  \doi{10.7937/K9/TCIA.2016.tNB1kqBU}

\bibitem{thompson1917growth}
Thompson, D.W.: {On Growth and Form}. Cambridge Univ. Press (1917)

\bibitem{yamao2020partial}
Yamao, K., Takenaka, M., Ishikawa, R., Okamoto, A., Yamazaki, T., Nakai, A.,
  Omoto, S., Kamata, K., Minaga, K., Matsumoto, I., et~al.: Partial pancreatic
  parenchymal atrophy is a new specific finding to diagnose small pancreatic
  cancer ($\leq$ 10 mm) including carcinoma in situ: comparison with localized
  benign main pancreatic duct stenosis patients. Diagnostics  \textbf{10}(7),
  ~445 (2020)

\bibitem{yang2017quicksilver}
Yang, X., Kwitt, R., Styner, M., Niethammer, M.: Quicksilver: {F}ast predictive
  image registration~-- {A} deep learning approach. NeuroImage  \textbf{158},
  378--396 (2017)

\bibitem{zhang2013bayesian}
Zhang, M., Singh, N., Fletcher, P.T.: Bayesian estimation of regularization and
  atlas building in diffeomorphic image registration. In: IPMI. vol.~23, pp.
  37--48 (2013)

\bibitem{Zimmerer2019UnsupervisedAL}
Zimmerer, D., Isensee, F., Petersen, J., Kohl, S.A.A., Maier-Hein, K.:
  Unsupervised anomaly localization using variational auto-encoders. In:
  MICCAI. pp. 289--297 (2019)

\end{thebibliography}

\end{document}


\title{Supplementary Material \\ Learning shape distributions from large databases of healthy organs: applications to zero-shot and few-shot abnormal pancreas detection}%

\titlerunning{Learning shape distributions from large databases of healthy organs}

\author{Rebeca V\'etil \and
Cl\'ement Abi-Nader \and
Alexandre B\^one \and
Marie-Pierre Vullierme \and
Marc-Michel Roh\'e \and
Pietro Gori \and
Isabelle Bloch
}

\authorrunning{V\'etil at al.}
\institute{}
\maketitle

\section{Model Architecture}
\begin{figure}[h]
\centering
\includegraphics[width=\textwidth]{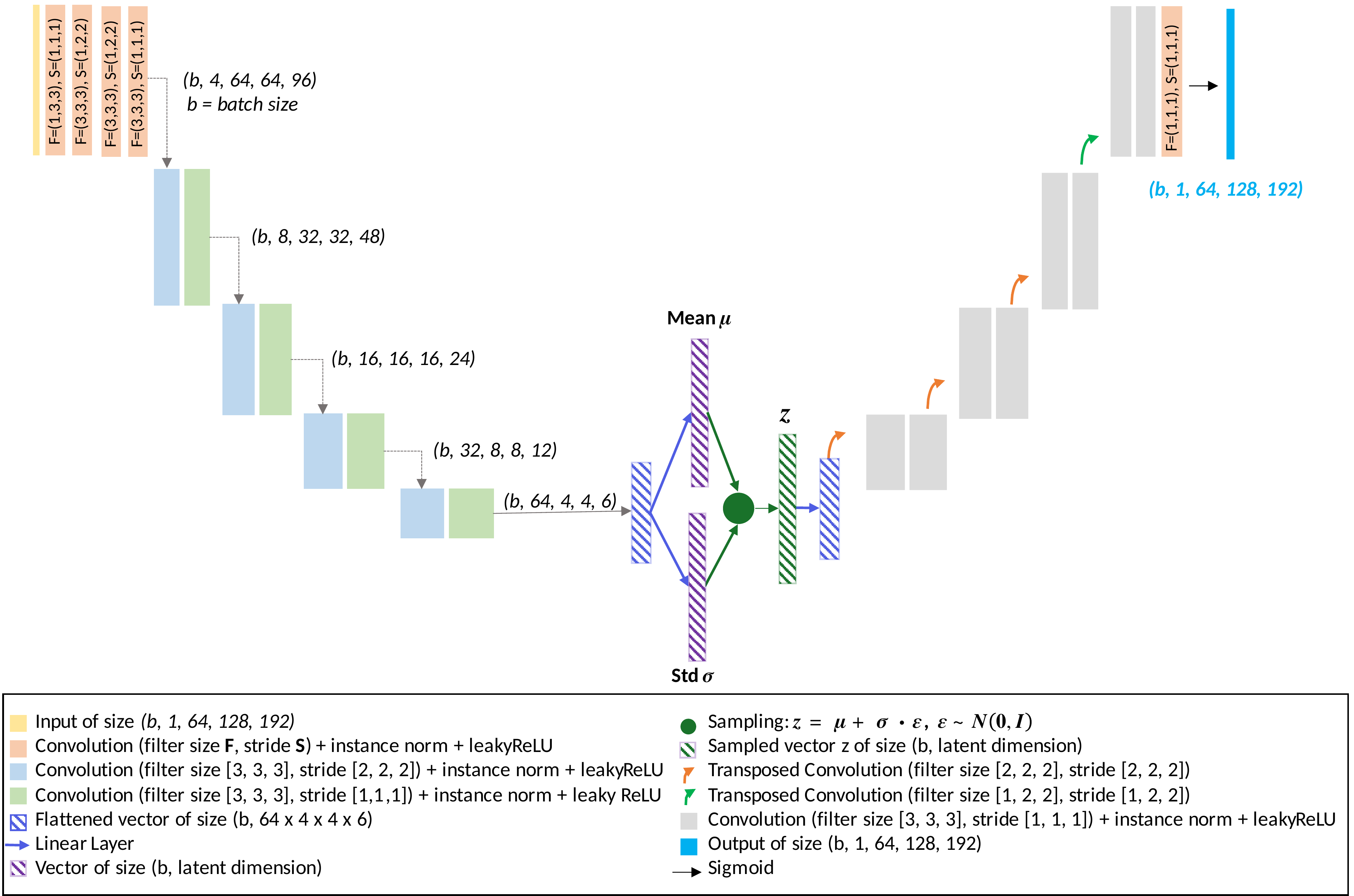}
\caption{\textbf{Architecture of the proposed VAE}} \label{fig:architecture}
\end{figure}

As detailed in Figure \ref{fig:architecture}, the proposed variational autoencoder (VAE) followed a 3D encoder-decoder architecture. The network topology (number of convolutions per block, filter sizes) was chosen based on the nnU-Net self-configuring procedure~\cite{isensee2021nnunet}. The model was trained on $800$ epochs. We used the Stochastic Gradient Descent optimizer with an initial learning rate of $10^{-4}$ following a polynomial decay. To avoid memory issues and mimic larger batch sizes, better suited for VAEs~\cite{kingma2014}, we used a batch size of $8$ with a gradient accumulation step of $5$.

\section{Inclusion criteria}
The training data were selected from a dataset consisting of $2606$ portal phase abdominal CT scans acquired from patients with potential liver disease. Each scan was accompanied by a radiological report as well as segmentations of the liver and liver lesions, done manually by radiologists. In order to exclude any patient whose condition could affect the pancreas, different exclusion criteria were applied. First, all patients with metastatic cancer or lesions reported as "unevaluable" were excluded. Secondly, in order to avoid overly large liver lesions, patients with lesion volumes exceeding the $99^{th}$ percentile were also excluded. Thirdly, cases in which lesions protruded from the liver were also removed from the study. Finally, the pancreas segmentation masks were automatically generated for the remaining patients, and only the cases for which the mask consisted of one single connected component were retained, this last criterion acting as a quality control on the segmentation masks. In the end, $1200$ cases were retained to build the healthy pancreas database for training.

\begin{table}[h]
\centering
\caption{Individual splits of the different training sets $\mathcal{D}_{N}$, as well as their corresponding inclusion criteria.}
\label{tab:splits}
\begin{tabular}{cccc||c|c|c||c}
\multicolumn{1}{c}{\multirow{2}{*}{\textbf{Name}}} & \multicolumn{1}{c}{\multirow{2}{*}{$N_{tot}$}} & \multicolumn{1}{c}{\multirow{2}{*}{$N_{train}$}} & \multicolumn{1}{c||}{\multirow{2}{*}{$N_{val}$}} & \multicolumn{3}{c||}{\textbf{Liver lesion}} & \multicolumn{1}{c}{\textbf{Other abdominal}} \\
\multicolumn{1}{c}{} & \multicolumn{1}{c}{} & \multicolumn{1}{c}{} & \multicolumn{1}{c||}{} & \multicolumn{1}{c|}{Benign} & \multicolumn{1}{c|}{Non suspicious} & \multicolumn{1}{c||}{Suspicious} & \multicolumn{1}{c}{\textbf{pathology}} \\ \hline
\textbf{\Dthree} & 300 & 240 & 60 & \cmark & \xmark & \xmark & \xmark \\
\textbf{\Dsix} & 600 & 480 & 120 & \cmark & \cmark & \xmark & \xmark \\
\textbf{\Dnine} & 900 & 720 & 180 & \cmark & \cmark & \cmark & \xmark \\
\textbf{\Dtwelve} & 1200 & 960 & 240 & \cmark & \cmark & \cmark & \cmark
\end{tabular}
\end{table}

\section{Reconstruction results}

As an additional indicator for the good reconstruction capacity of the proposed model, we also reported, in Table \ref{tab:dices}, the Dice scores computed between the input and output of the VAE for the $244$ subjects of $\mathcal{D}^{Test}$. As for the AUC scores obtained in zero-shot AD, the Dice scores increased with larger latent space dimensions and larger training sets. In particular, best reconstruction performances coincided with the best detection performances, obtained on the model trained on $\mathcal{D}_{1200}$ with a latent dimension $L = 1024$.
\begin{table}[h]
\centering
\caption{\textbf{Results for segmentation masks reconstruction.} For each experiment, corresponding to a specific training size and latent space dimension, we report the mean and standard deviation of Dice scores obtained by bootstrapping with $10000$ repetitions. Best results by line are \underline{underlined} and by column are in \textbf{bold}.} 

\label{tab:dices}
\begin{tabular}{c||c|c|c|c}
 & $L_{16}$ & $L_{64}$ & $L_{256}$ & $L_{1024}$ \\ \hline
\textbf{\Dthree} & 73.7 \scriptsize{$\pm$7.2} & 79.9 \scriptsize{$\pm$5.7} & 80.9 \scriptsize{$\pm$7.4} & \underline{81.5 \scriptsize{$\pm$7.4}} \\ \hline
\textbf{\Dsix} & 75.5 \scriptsize{$\pm$7.5} & 83.4 \scriptsize{$\pm$5.4} & \underline{85.0 \scriptsize{$\pm$5.8}} & 83.9 \scriptsize{$\pm$5.9} \\ \hline
\textbf{\Dnine} & \textbf{78.1 \scriptsize{$\pm$5.9}} & 85.0 \scriptsize{$\pm$5.8} & \underline{86.1 \scriptsize{$\pm$5.4}} & 85.4 \scriptsize{$\pm$5.8} \\ \hline
\textbf{\Dtwelve} & 77.3 \scriptsize{$\pm$6.4} & \textbf{85.9 \scriptsize{$\pm$5.4}} & \textbf{86.2 \scriptsize{$\pm$5.3}} & \underline{\textbf{88.5 \scriptsize{$\pm$4.4}}} \\ \hline
\end{tabular}%
\end{table}

\section{Comparison with other SSM methods}
We compared our framework with two state-of-the-art methods: active shape models (ASM)~\cite{cootes1995_ASM} and Large Deformation Diffeomorphic Metric Mapping (LDDMM) using the Deformetrica software~\cite{bone2018_deformetrica}. Concerning ASM, we computed the signed distance map of the pancreas 3D contours of each subject. For LDDMM, we estimated a Bayesian Atlas \cite{gori2017bayesian} parametrized by $576$ control points. In both cases, we performed a PCA on the shape-encoding parameters to obtain a latent vector of dimension $1024$ for each subject. We compared them with our method also using a latent dimension $L = 1024$. All methods were trained on $\mathcal{D}_{1200}$. Results, reported in Table \ref{tab:asm_lddmm}, show that our method outperforms ASM and LDMM when the number of training samples is small (Train/Test ratio $\leq 0.05$).

\begin{table}[h]
\centering
\caption{\textbf{Few-shot and zero-shot AD results.} For each experiment, we report the mean and standard deviation for AUC (in \%), obtained by bootstrapping with $10000$ repetitions. Best results by column are in \textbf{bold}.} 

\label{tab:asm_lddmm}
\begin{tabular}{c||c||ccc}
\textbf{Configuration} & \cellcolor[HTML]{343434}{\color[HTML]{FFFFFF} \textbf{Zero-shot}} & \multicolumn{3}{c}{\cellcolor[HTML]{343434}{\color[HTML]{FFFFFF} \textbf{Few-shot}}}                                                                                                                          \\
\textbf{Train/Test ratio}                          & \textbf{0}         & \multicolumn{1}{c|}{\textbf{0.05}} & \multicolumn{1}{c|}{\textbf{1}}   & \textbf{223} \\ \hline
\textbf{Ours}             & \textbf{65.41\scriptsize{$\pm$0.36}}     & \multicolumn{1}{c|}{\textbf{66.02\scriptsize{$\pm$0.02}}}   & \multicolumn{1}{c|}{86.95\scriptsize{$\pm$0.23}} & 91.18\scriptsize{$\pm$0.19} \\
\textbf{LDDMM}            & 54.6\scriptsize{$\pm$0.36}        & \multicolumn{1}{c|}{58.68\scriptsize{$\pm$0.02}}  & \multicolumn{1}{c|}{\textbf{89.43\scriptsize{$\pm$0.19}}} & \textbf{95.41\scriptsize{$\pm$0.11}} \\
\textbf{ASM}              & 58.42\scriptsize{$\pm$0.36}       & \multicolumn{1}{c|}{61.1\scriptsize{$\pm$0.02}}  & \multicolumn{1}{c|}{82.64\scriptsize{$\pm$0.26}} & 93.79\scriptsize{$\pm$0.14}
\end{tabular}%
\end{table}

\bibliographystyle{splncs04}
\bibliography{biblio}
